\documentclass[conference]{IEEEtran}
\IEEEoverridecommandlockouts
\usepackage{cite}
\usepackage{amsmath,amssymb,amsfonts}
\usepackage{algorithmic}
\usepackage{graphicx}
\usepackage{textcomp}
\usepackage{multirow}
\usepackage{xcolor}
\usepackage{url}
\usepackage{comment}
\usepackage{enumitem}
\usepackage{colortbl}
\usepackage[colorlinks=true, urlcolor=blue, linkcolor=red]{hyperref}
\DeclareMathOperator*{\argmax}{arg\,max}
\DeclareMathOperator*{\argmin}{arg\,min}
\usepackage{mathtools}
\usepackage{algorithmic}
\usepackage{algorithm}
\usepackage{array}
\usepackage{booktabs} 
\def\BibTeX{{\rm B\kern-.05em{\sc i\kern-.025em b}\kern-.08em
    T\kern-.1667em\lower.7ex\hbox{E}\kern-.125emX}}
\begin{document}

\twocolumn[{%
\vspace{20mm}
{ \large
\begin{itemize}[leftmargin=2.5cm, align=parleft, labelsep=2cm, itemsep=4ex,]

\item[\textbf{Citation}]{K. Kokilepersaud*, Y. Logan*, R. Benkert, C. Zhou, M. Prabhushankar, G. AlRegib, E. Corona, K. Singh, A. Parchami, "FOCAL: A Cost-Aware, Video Dataset for Active Learning," in \textit{IEEE International Conference on Big Data ,} Sorento, Italy, Dec. 15-18, 2023}

\item[\textbf{Review}]{Date of Publication: December 15th 2023}

\item[\textbf{Codes}]{\url{https://github.com/olivesgatech/FOCAL_Dataset}}

\item[\textbf{Data}]{\url{https://zenodo.org/records/10145325}}

\item[\textbf{Bib}]  {@inproceedings\{kokilepersaud2023focal,\\
    title=\{FOCAL: A Cost-Aware, Video Dataset for Active Learning\},\\
    author=\{Kokilepersaud, Kiran and Logan, Yash-Yee and Benkert, Ryan and Zhou, Chen and Prabhushankar, Mohit and AlRegib, Ghassan and Corona, Enrique and Singh, Kunjan and Parchami, Armin\},\\
    booktitle=\{IEEE International Conference on Big Data\},\\
    year=\{2023\}\}}


\item[\textbf{Contact}]{
\{kpk6, mohit.p, alregib\}@gatech.edu\\\url{https://ghassanalregib.info/}\\}
\end{itemize}

}}]

\newpage

\title{FOCAL: A Cost-Aware Video Dataset for Active Learning\\

\thanks{$^{\diamond}$ = equal contribution}
}

\author{\IEEEauthorblockN{Kiran Kokilepersaud$^{\diamond}$$^{\star}$, Yash-Yee Logan$^{\diamond}$$^{\star}$, Ryan Benkert$^{\star}$, Chen Zhou$^{\star}$, Mohit Prabhushankar$^{\star}$, \\ Ghassan AlRegib$^{\star}$, Enrique Corona$^{\dagger}$, Kunjan Singh$^{\dagger}$, Mostafa Parchami$^{\dagger}$}
\IEEEauthorblockA{\textit{$^{\star}$ OLIVES at the Center for Signal and Information Processing CSIP,} \\
\textit{ $^{\star}$ School of Electrical and Computer Engineering, Georgia Institute of Technology, Atlanta, GA, USA}\\
\textit{$^{\dagger}$ Ford Motor Company, Dearborn, Michigan}\\
\{kpk6,logany,rbenkert3,chen.zhou,mohit.p,alregib\}@gatech.edu \{ecoron18,ksing114,mparcham\}@ford.com}

}

\maketitle
\begin{abstract}
In this paper, we introduce the \texttt{FOCAL} (Ford-OLIVES Collaboration on Active Learning) dataset which enables the study of the impact of annotation-cost within a video active learning setting. Annotation-cost refers to the time it takes an annotator to label and quality-assure a given video sequence. A practical motivation for active learning research is to minimize annotation-cost by selectively labeling informative samples that will maximize performance within a given budget constraint. However, previous work in video active learning lacks real-time annotation labels for accurately assessing cost minimization and instead operates under the assumption that annotation-cost scales linearly with the amount of data to annotate. This assumption does not take into account a variety of real-world confounding factors that contribute to  a nonlinear cost such as the effect of an assistive labeling tool and the variety of interactions within a scene such as occluded objects, weather, and motion of objects. \texttt{FOCAL} addresses this discrepancy by providing real annotation-cost labels for 126 video sequences across 69 unique city
scenes with a variety of weather,
lighting, and seasonal conditions. These videos have a wide range of interactions that are at the intersection of infrastructure-assisted autonomy and autonomous vehicle communities. We show through a statistical analysis of the \texttt{FOCAL} dataset that cost is more correlated with a variety of factors beyond just the length of a video sequence. We also introduce a set of conformal active learning algorithms that take advantage of the sequential structure of video data in order to achieve a better trade-off between annotation-cost and performance while also reducing floating point operations (FLOPS) overhead by at least 77.67\%. We show how these approaches better reflect how annotations on videos are done in practice through a sequence selection framework. We further demonstrate the advantage of these approaches by introducing two performance-cost metrics and show that the best conformal active learning method is cheaper than the best traditional active learning method by 113 hours.  The code associated with this paper can be found at \href{https://github.com/olivesgatech/FOCAL_Dataset}{this repository}. The data can be downloaded at \href{https://zenodo.org/records/10145325}{this location}.
\end{abstract}


\section{Introduction}

\begin{figure}[htb!]
\centering
\includegraphics[scale=.22]{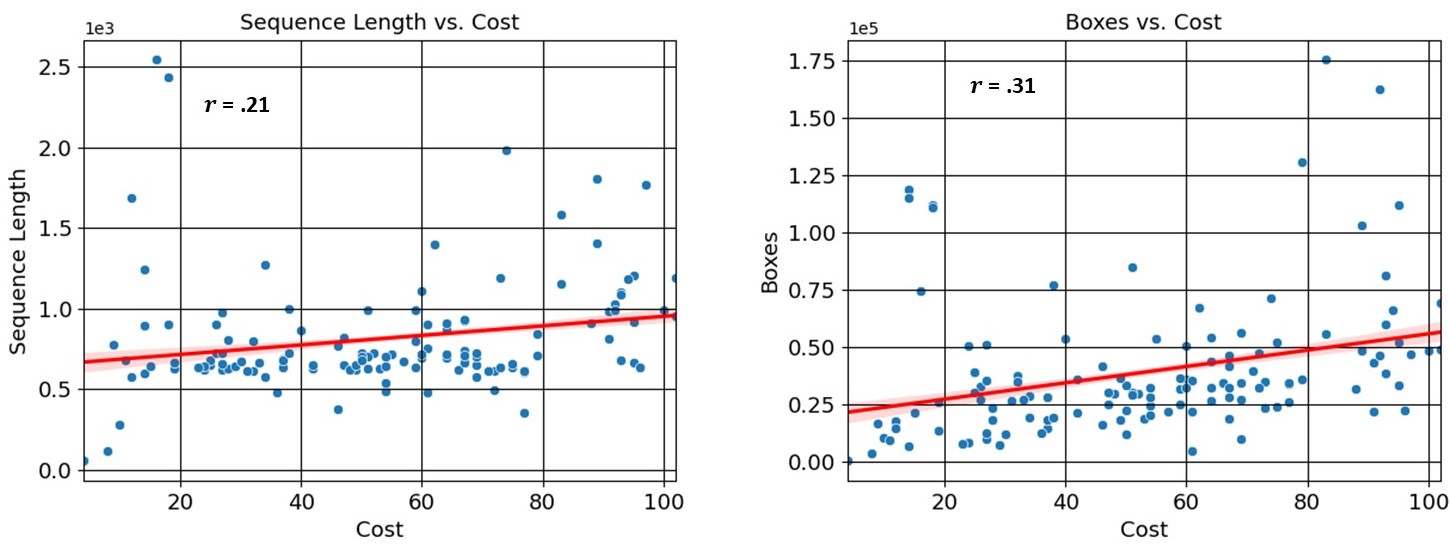}
\caption{This shows a plot of the cost vs. sequence length and cost vs. box counts for every video in the \texttt{FOCAL} dataset. The pearson correlation coeffcient between these variables is .21 and .31 respectively.} \vspace{-6mm}
\label{fig: sequence_length_cost}
\end{figure}

\begin{figure*}[htb!]
\centering
\includegraphics[scale=.35]{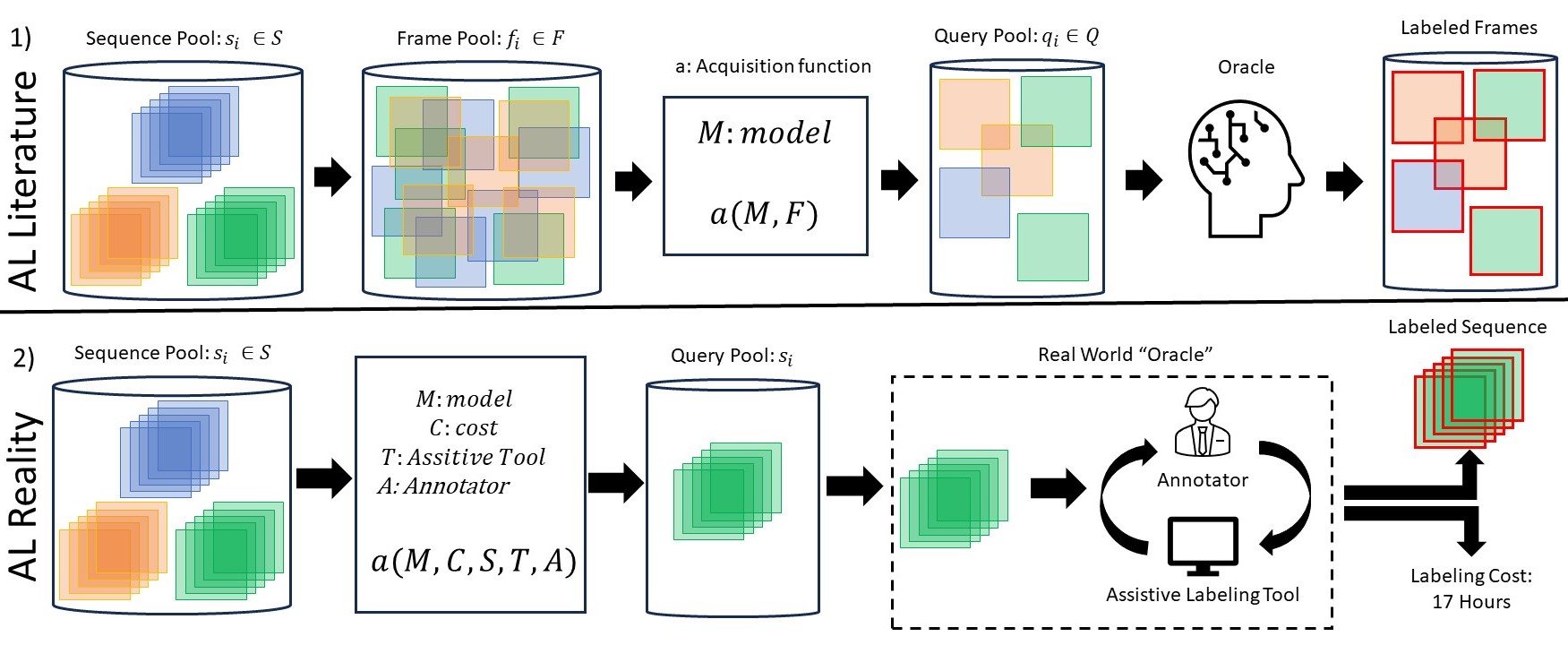}
\caption{1) This shows how traditional active learning algorithms processes sequences. All frames of a sequence are equally likely to be selected and labels can be extracted from an Oracle. 2) This shows how in a real world context, annotators are assigned sequences to select to label and then use assistive labeling tools to interpolate between fully labeled frames. This results in both a fully labeled sequence and an associated cost for the work done.}
\label{fig: real_trad_AL}
\end{figure*}

Machine learning has demonstrated the capability to have a major impact in a variety of application areas such as health-care \cite{kokilepersaud2023clinically}, seismology \cite{alaudah2019machine}, and autonomous driving \cite{kokilepersaud2023exploiting}. Despite the potential that machine learning presents, the efficacy of these algorithms is dependent on access to an abundance of high quality labeled data related to the application of interest. This dependence introduces a host of issues because annotating data is an expensive and time-consuming process \cite{fredriksson2020data}. This problem is especially prevalent in domains that involve videos \cite{jiao2021new} where long sequences of spatio-temporally correlated frames complicate labeling from both a cost and complexity point of view. In practice, annotators address these issues by making use of assistive labeling technology \cite{ashktorab2021ai} that reduces annotator burden by interpolating labels across correlated frames in a video sequence to produce a fully labeled sequence in an efficient manner. Due to the advantages these tools provide, labeling is typically done across sequential blocks, rather than on single frames at a time \cite{yuen2009labelme}. Alongside the tools utilized, another practical concern of annotators is how to decide which video sequences to label. Ideally, annotators will choose the sequences that result in the least annotation-cost while maintaining high downstream performance in the video task of interest. We define annotation-cost as the amount of time it takes an annotator to label and quality-assure their assigned labeling task. 

The study of this interaction between annotation-cost and downstream performance is known as active learning \cite{settles2009active}. The motivation behind active learning is to design query selection strategies that select informative samples for annotators to label. These selected samples ideally maximize performance within a given budget constraint. Despite this motivation related to reducing cost, traditional active learning algorithms do not have access to the exact annotation-cost when benchmarking their algorithms and make the assumption that cost shares a linear relationship with the number of samples labeled \cite{settles2008activecost}. However, this assumption of a linear
relationship between cost and samples is not valid across
application domains such as video. A variety of other factors beyond just the amount of data can affect annotation-cost. This includes time-saved from using an assistive labeling tool and the complexity of interactions within the data itself such as motion, object density, occlusion severity, and lighting conditions.  We show in Figure \ref{fig: sequence_length_cost} a plot between annotation-cost of a video and its associated number of frames and bounding box counts. This figure demonstrates only a minor correlation with a Pearson correlation coefficient of .21 for frames and .31 for box counts which is far from the linear correlation suggested by active learning literature. 

Another issue with traditional active learning strategies is that they do not reflect how video data is annotated in practice. Many query strategies such as \cite{wang2014new,ash2019deep,benkert2022gauss} were designed on classification tasks that do not take into account spatio-temporal relationships that exist in video data. This is shown in part 1) of Figure \ref{fig: real_trad_AL} where the sequence pool $S$ is decomposed to a set of frames $f_i \in F$ from which an acquisition function $a$ takes as input the current model $M$ and the pool of frames $F$ to sample from. The chosen frames are then labeled by an oracle that provides the labels for the selected data. However, as discussed previously, the presence of assistive labeling technology enables annotating in a sequence-wise manner. This real-world sequential labeling process is demonstrated in part 2) of Figure \ref{fig: real_trad_AL} which results in a labeled sequence with an associated annotation-cost from the time spent by the worker. Additionally, it should be noted that the acquisition of new sequences is dependent on a variety of factors not accounted for in the active learning literature. This includes the model $M$, potential cost of the sequence $C$, the impact of factors in a video sequence that effect the assistive labeling tool $T$, and the usage of the tool by the annotator $A$. This discrepancy between real-world practice and standard frameworks suggest that active learning algorithms should be tested in settings that reflect actual annotator practice and that query strategies should be designed that reflect characteristics of the sequential structure of the video domain.

\begin{table*}[t!]
\tiny
\centering
\caption{A comparisons of datasets used in cost-aware active learning.}
\label{tab:dataset_comparisons}
\scalebox{1.3}{
\begin{tabular}{clccccc} 
\toprule
\textbf{Domain}     & \multicolumn{1}{c}{\textbf{Dataset}}                            & \textbf{Cost} & \textbf{Cost Unit} & \textbf{Cost Added}    & \textbf{Method}    & \textbf{Size}                    \\ 
\hline \midrule
\multirow{11}{*}{NLP}  & \multirow{2}{*}{Wall Street Journal Corpus \cite{marcinkiewicz1994building}} & Approx.       & \# Brackets        & Prospectively          & Sequences          & \multirow{2}{*}{2350 Sentences}  \\
                       &                                             & Real          & Time               & Retrospectively        & Sequences          &                                  \\
                       & Hotel Reviews \cite{KaggleDatafiniti}                               & Approx.       & Samples            & -                      & Sequences          & 10000 Reviews                    \\
                       & IMDB \cite{maas2011learning}                                       & Approx.       & Samples            & -                      & Sequences          & 50000 Reviews                    \\
                       & Review Polarity  \cite{pang2004sentimental}                           & Approx.       & Samples            & -                      & Sequences          & 2000 Reviews                     \\
                       & Sentence Polarity \cite{pang2005seeing}                          & Approx.       & Samples            & -                      & Sequences          & 10662 Reviews                    \\
                       & Wikipedia Movie Plots \cite{KaggleWikipedea}                     & Approx.       & Samples            & -                      & Sequences          & 33869 Descriptions               \\
                       & CKB News Corpus \cite{settles2008activecost}                             & Real          & Time               & Retrospectively        & Sequences          & 1984 Articles                    \\
                       & Speculative Text Corpus \cite{settles2008activecost}                    & Real          & Time               & Retrospectively        & Sequences          & 850 Sentences                    \\
                       & SigIE Email Corpus \cite{settles2008activecost}                         & Real          & Time               & Retrospectively        & Sequences          & 250 Signatures                   \\
                       & Urdu-English Language Pack \cite{LDC}                 & Real          & Time  Money        & Retrospectively        & Sequences          & 88000 Sentences                  \\ 
\midrule
\multirow{2}{*}{Image} & SIVAL Image Repository  \cite{rahmani2006missl}                    & Real          & Time               & Retrospectively        & Frames             & 1500 Images                      \\
                       & Minimum Required Viewing Time \cite{mayohard} ~              & Real          & Time               & Retrospectively        & Frames             & 4771 Images                      \\ 
\midrule
\multirow{2}{*}{Video} & VIRAT \cite{oh2011large} ~                                      & Approx.       & Samples            & -                      & Frames             & 16 videos (2700000 frames)                       \\
                       & \textbf{FOCAL}                              & \textbf{Real} & \textbf{Time}      & \textbf{Prospectively} & \textbf{Sequences} & \textbf{126 Videos (103518 frames)}              \\
\bottomrule
\end{tabular}}
\end{table*}
In this paper, we address both of these issues of conventional active learning through the introduction of the \texttt{FOCAL} dataset. This is the first video active learning dataset with associated annotation-cost labels for each sequence that was recorded during the labeling of the dataset. These videos took place in practical settings that are of interest to both the autonomous vehicle and infrastructure-assisted autonomy communities. Through this dataset, we benchmark active learning algorithms with respect to performance and the cost associated with the queried sequences of each algorithm. Additionally, through an analysis of the dataset we identify factors that contribute to annotation-cost and use these factors to design conformal query strategies. Conformal refers to algorithms that ``conform" or make use of the data structure to devise importance scores for sampling. These strategies take advantage of temporal and spatial statistics of video to minimize cost while maximizing performance. We also introduce two new metrics that reflect the interaction between cost and performance. This is done within an active learning experimental setting that reflects real world annotation workflows where entire sequences are selected rather than individual frames. In summary, the contributions of this paper are as follows:

\begin{itemize}
\item We introduce the first public video active learning dataset accompanied by detailed annotations and annotation time, that provides a rigorous basis for accurate quantitative evaluation of annotation cost reduction in active learning research. 
\item We conduct a detailed analysis of multiple factors that affect the annotation cost of video sequences. 
\item We provide a new benchmark for object detection and cost-conscious video active learning on \texttt{FOCAL}. 
\item We compare adaptions of traditional inferential sampling to conformal sampling on video sequences.
\item We introduce two new performance-cost metrics for validating the effectiveness of query strategies minimizing cost while maximizing performance.
\item The presence of video statistics allows us to front-load ranking of the unlabeled pool, thus reducing overhead cost by greater than 8 million GFLOPS.
\end{itemize}

\section{Related Works}

\subsection{Cost Measurement} Cost is defined and measured differently across several domains of cost-conscious active learning. In the domain of natural language processing, \cite{hwa2000sample} propose to measure cost using the number of brackets within sentences. Other works use the number of sentences and the number of decisions needed to select a sentence parse as metrics to measure cost in active learning \cite{osborne2004ensemble}.\cite{tomanek2009semi} measures cost by the number of tokens for parts of sentences while \cite{kapoor2007selective} simulates cost to be a linear function of the length of a voicemail message. These are examples of approximate costs being used for active learning. In contrast, \cite{ngai2001rule} measured actual annotation time to compare the efficacy of rule writing versus annotation with active learning. Overall, \cite{baldridge2009well} considered actual annotation time to be the best metric for cost measurement due to the variation that results from the practices employed by annotators in real labeling settings. We are the first to provide these measurements for the video domain by defining cost as the time to annotate and quality assure bounding box labels in a full sequence.

\begin{figure*}[htb!]
\centering
\includegraphics[scale=.40]{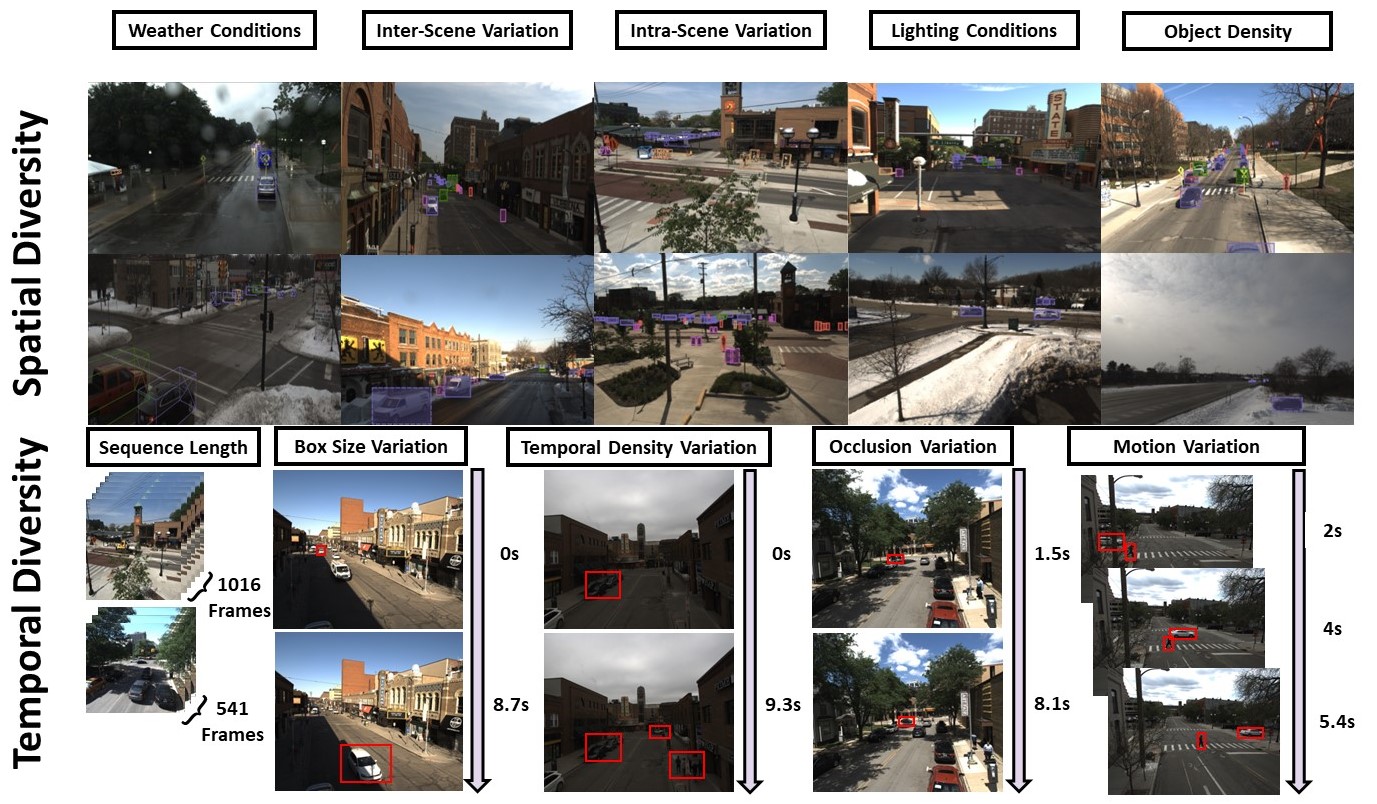}
\caption{Examples of Spatial and Temporal Diversity throughout the \texttt{FOCAL} dataset.}
\label{fig: space_temp_diversity}
\end{figure*}

\subsection{Cost-Available Datasets} We summarize several key characteristics of our FOCAL dataset and other datasets used for cost-conscious active learning in Table \ref{tab:dataset_comparisons}. These works demonstrate how active learning strategies differ based on practical considerations surrounding relevant application settings. The works where approximate cost is reduced by querying fewer samples do not actually represent or reason about cost in practical settings. The studies that use real costs attempt to provide insight on the role of annotation cost in real-world active learning. However, those studies used datasets that did not originally come with annotation cost labels. The authors retrospectively used tools such as Amazon Mechanical Turk to gather the annotations. Retrospective annotations are known to have more sources of confounding error and bias \cite{gerhard2008bias}. \texttt{FOCAL} dataset bridges this gap between simulated and real-world active learning considerations and serves to provide a more robust means of benchmarking active learning algorithms. To the best of our knowledge, \texttt{FOCAL} is the first video dataset intentionally designed for sequential active learning with real time \textit{prospective cost annotations}.

\subsection{Cost-Conscious Sampling} Cost-conscious sampling refers to previous work that attempted to integrate cost into  query strategies. For example, \cite{hu2019active} designed a query strategy that uses partial labels to reduce cost. \cite{biswas2012active} developed queries with clustering algorithms to remove the need to annotate samples belonging to an already annotated cluster. \cite{qian2013active} developed a similarity-based query that rank the importance of a sample's neighbors such that less time is needed to label each sample. \cite{vondrick2011video} used a video active learning strategy based on selecting identified frames for a video tracking task. However, sampling a subset of frames in this manner does not match the human annotation workflow for video data and imposes an impractical method of annotating video data. Also, like many other studies, this method considers cost as constant between samples and only reduces it by querying less frames.

\subsection{Active Learning} Traditional active learning methods typically involve a model iteratively selecting informative samples for annotation from an unlabeled data pool until an annotation budget is reached. Approaches for selecting samples differ in their definition of information content. For example, several approaches define sample importance using softmax probabilities \cite{wang2014new, roth2006margin} where information content is related to the output logits of the network. Other works focus on constructing the core-set of the unlabeled data pool \cite{sener2017active, longtailcoreset}. Furthermore, there are several approaches that consider the combination of both data representation and generalization difficulty within their definition of information content \cite{ash2019deep, prabhushankar2022introspective, haussmann2020scalable}. \cite{batchbald} integrates both generalization difficulty and data representation by extending \cite{houlsby2011bayesian} to diverse batch acquisitions. Recent work \cite{benkert2022gauss} defined sample importance in relation to the number of times a sample changes its output prediction. 
There is also a large body of active learning work that studies data collection within specific application domains. \cite{logan2022decal} and \cite{logan2022patient} analyze how clinical information can be incorporated into active learning query strategies. \cite{mustafa2021man} utilizes auto-encoder reconstruction information to better select samples from seismic volumes. \cite{benkert2022false} introduces the idea of analyzing how out of distribution scenarios manifest in active learning. Active learning has also been applied within the video domain. \cite{bengar2019temporal} made use of temporal coherence information as a means for querying samples. \cite{rana2022all} identified specific frames to label within a video action detection task. Although these works are motivated by the practical goal of active learning - reduce labeling costs, while maintaining a high performance - they each operate on the assumption that annotation costs are constant between instances. The authors in~\cite{settles2008activecost} state that most publicly available datasets are not created with the intention of reducing annotation cost via active learning research. This means that annotation time, or any other form of cost, was not logged for each instance. Our work bridges the gap through the introduction of a dataset with explicit cost labels within a video setting. Additionally, we use a novel experimental setup that utilizes the realistic sequence-wise acquisition paradigm.

\section{FOCAL Dataset}

\begin{table*}[]
\small
\centering
\caption{This table shows computed correlation coefficient scores between the cost and each of the listed statistics below. Most costly refers to the 20 sequences that took the most time to label. Least costly refers to the 20 sequences that took the least time to label. P,K, and S refer to the pearson, kendall, and spearman correlation coefficients.}
\label{tab:correlation}
\begin{tabular}{@{}cccccccccc@{}}
\toprule
\multicolumn{10}{c}{FOCAL Dataset Statistical Correlation with Cost}                 \\ \midrule
Statistic & \multicolumn{3}{c}{Total Dataset} & \multicolumn{3}{c}{Most Costly} & \multicolumn{3}{c}{Least Costly} \\
                   & P    & K    & S    & P     & K     & S     & P    & K    & S    \\ \midrule
Sequence Length    & 0.21 & 0.22 & 0.31 & -0.27 & -0.10 & -0.17 & 0.26 & 0.25 & 0.34 \\
Number of Objects  & 0.31 & 0.34 & 0.46 & -0.37 & -0.03 & -0.05 & 0.25 & 0.26 & 0.35 \\
Occlusion Severity & 0.25 & 0.26 & 0.37 & 0.05  & 0.021 & 0.12  & 0.14 & 0.18 & 0.23 \\
Motion             & 0.21 & 0.17 & 0.24 & -0.14 & 0.00  & -0.04 & 0.14 & 0.06 & 0.12 \\
Season             & 0.07 & 0.07 & 0.09 & 0.24  & 0.20  & 0.25  & 0.17 & 0.08 & 0.12 \\
Time of Day        & 0.11 & 0.07 & 0.10 & 0.17  & 0.15  & 0.21  & 0.15 & 0.15 & 0.26 \\ 
Number of Cars        & 0.17 & 0.21 & 0.30 & -0.48  & -0.04  & -0.02  & 0.12 & 0.26 & 0.36 \\ 
Number of Pedestrians        & 0.26 & 0.28 & 0.39 & -0.15  & -0.17  & -0.23  & 0.03 & 0.08 & 0.16 \\ 
\bottomrule
\end{tabular}
\end{table*}

\subsection{Dataset Construction}
The \texttt{FOCAL} dataset is composed of 126 video sequences across 69 unique scenes. These scenes were collected from a variety of urban locations across Ann Arbor and Detroit Michigan from January 2021 to August 2021. Videos are taken at diverse points in the year as well as times of day to increase spatial diversity. This includes variation with respect to weather conditions, lighting conditions, object density, objects within scenes, and objects between scenes. The data collection locations of these video sequences
were also selected based on the intent to obtain a wide range
of interactions that are at the intersection of infrastructure-
assisted autonomy and autonomous vehicle communities. This includes a variety of interactions between pedestrians and vehicles in traffic settings. Additionally, the collected scenes exhibit a wide range of temporal diversity  which includes variation in sequence length, object size, number of moving objects, number of hidden objects, and velocity of moving objects. These desirable characteristics are visually displayed in Figure \ref{fig: space_temp_diversity}.

\begin{figure*}[htb!]
\centering
\includegraphics[scale=.65]{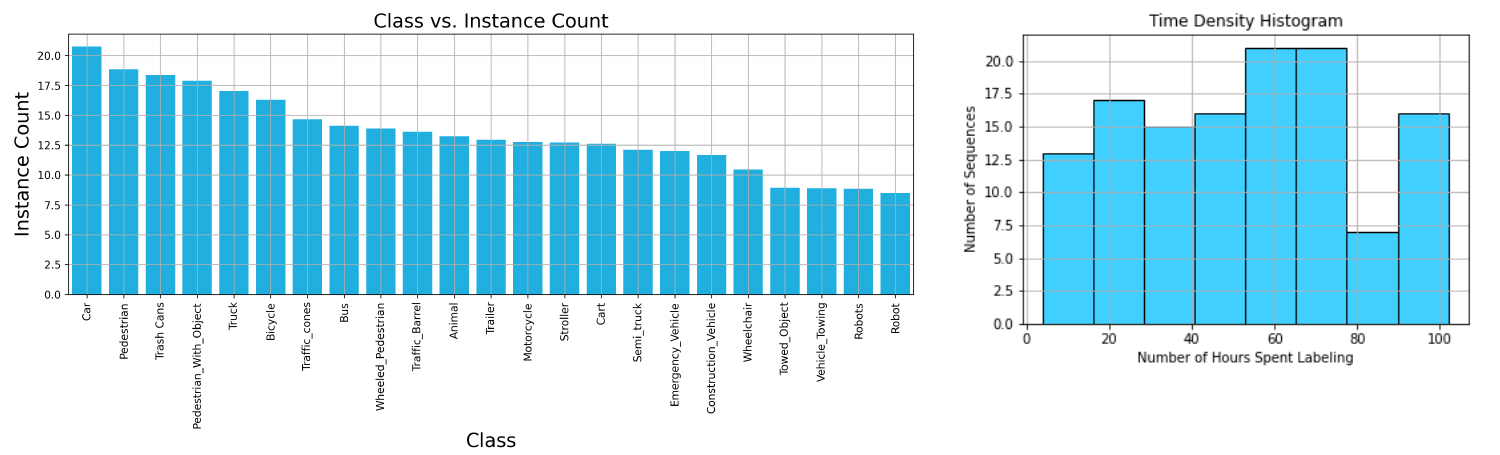}
\caption{FOCAL dataset class distribution and variation of cost annotations. Left: Log-scaled class distribution. The y-axis shows the number
of object instances for each class in the FOCAL dataset. Right: Variation of cost within frames in sequence. The variation is beneficial for
distinguishing active learning algorithms to identify optimal strategies that simultaneously select sequences with less time to annotate and
optimize for performance.} \vspace{-.3cm}
\label{fig: time_histogram}
\end{figure*}

During the labeling process, annotators are assigned video sequences within an assistive labeling tool platform. They assigned bounding box labels to objects with respect to 23 possible fine-grained classes. The distribution of these classes as a function of object counts is shown in Figure \ref{fig: time_histogram}. We note that there is a relatively uniform distribution across the majority of object categories.  In addition to bounding boxes, annotators also label objects with respect to a variety of different types of metadata. This includes tracking identities and category specific attributes. These attributes include the occlusion status of an object, whether car doors are open or closed, the parking status of cars, and whether pedestrians are walking or standing still.  We also show other statistics of interest in Table \ref{tab:stats}. Namely, we show a representative spread with respect to different seasons, the number of objects/frame (label density), and the lengths of different sequences. We choose sequences with a relatively even number of frames in order to avoid potential bias related to sequence length disproportionately influencing results. Sequences were also chosen to have a wide variation in label density to allow active learning strategies to query from a diverse sequence pool.

\begin{table}[]
\centering
\caption{General Statistics from \texttt{FOCAL} dataset.}
\label{tab:stats}
\begin{tabular}{@{}cc@{}}
\toprule
\multicolumn{2}{c}{Focal Dataset Statistics} \\ \midrule
Number of Classes                & 23          \\
Number of SuperClasses           &   4        \\
Seasonal Distribution            &   10.3\% Winter / 26.2\% Summer / 63.5\% Spring        \\
Label Density                    &   $47.11 \pm 26.96$        \\
Sequence Lengths                 &  $812 \pm 368$         \\
Train/Test/Val Split             &  51 Scenes / 13 Scenes / 5  Scenes        \\ \bottomrule
\end{tabular}

\end{table}

\subsection{Cost Analysis}
During the labeling process, annotators were timed by their assistive labeling platform to arrive at a final annotation-cost label for each sequence. It consists of the time spent in the platform annotating bounding boxes and attributes, the time spent for the sequence to undergo customer quality assurance, and the time to perform final quality assurance. This is all computed with respect to hours. We show the distribution of annotation-cost across all sequences in the dataset in Figure \ref{fig: time_histogram}. We note that there are a relatively equal number of sequences for different ranges of annotation-cost. This presents a wide pool of different cost sequences for active learning algorithms to select from. 

We also analyze what factors cause certain sequences to take longer to label than others. Intuitively the length of a
sequence should be proportional to the annotation cost. However, Figure \ref{fig: sequence_length_cost} shows a minor correlation between cost and sequence length in the \texttt{FOCAL} dataset. A long sequence may or may not contain
many relevant objects to be annotated and the same is true for
a short sequence. We explore potential other factors that can effect cost in Table \ref{tab:correlation}. We compute the pearson, kendall, and spearman correlation coeffcients between the annotation-cost and a variety of different statistics in the dataset. Additionally, we compute these metrics over different subsets of the dataset which includes the total dataset, the 20 most costly sequences, and the 20 least costly sequences. The first point that stands out is that a variety of other factors have a positive correlation with cost. The most intuitive correlations are the number of boxes and the number of occluded objects in a scene. More boxes indicates more time spent labeling and occluded objects are more difficult to label because a portion of the object is hidden behind part of the scene. The presence of more cars and pedestrians can also complicates labeling by having more moving objects with diverse interactions. Additionally, object motion as computed by optical flow on \cite{sun2018pwc}, the season, and the time of day also influence how easily different objects can be identified during the annotation process and exhibit some correlation with cost. However, it is interesting to note the discrepancy when we restrict the metric computation to the most and least costly sequences. The least costly sequences are the easiest to label and have more straightforward patterns that are easily identified by annotators. For this reason, they exhibit positive correlations with all the statistics of interest. This trend does not hold for the most costly sequences. Intuitive metrics such as number of frames and objects now either have negligible or negative correlation with cost. This suggests that higher cost sequences take longer to label not because of our traditional understanding of cost, but as a result of more complex interactions in the scene. As a result, it only exhibits a positive cost correlation with the season, time of day, and the occlusion severity which are statistics that are more reflective of the kinds of interactions happening in the scene. We make use of insights from this analysis in the design of our conformal sampling strategies.


\section{Experimental Setup}

\begin{figure} [t!]
\centering
\includegraphics[width=0.85\columnwidth]{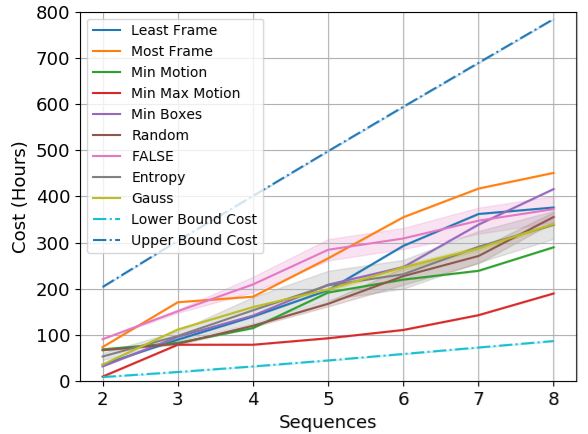}
\caption{We compare annotation cost, in hours, of all sampling algorithms to the theoretical upper and lower cost bounds of the \texttt{FOCAL} dataset. Shaded curves show the average cost and standard error for algorithms with stochastic cost.} \vspace{-7mm}
\label{fig:upper-lower-cost-bounds}
\end{figure} 

\subsection{Active learning Experimental Setup}
All experiments are conducted with
a YOLOv5n \cite{Jocher_YOLOv5_by_Ultralytics_2020} architecture. All object detection experiments took place with an ADAM optimizer, a learning rate  of .01, a weight decay of .00005, momentum of .937, and a step scheduler that decreases by a factor of 10 after every 3 epochs. The batch size is set at 16, the iou threshold is .6, and the confidence threshold for non maximum suppression is set at .65. All images are resized to $640\times640\times3$ and the training augmentations include horizontal flipping, translations, scaling, color jitter, and mosaic grids. Within each active learning round,
the model is trained for 10 epochs. After each epoch, the best
model with respect to the validation set is saved to evaluate
the test set for that round. This is repeated with three different
random seeds and mean average precision(mAP) was computed. 

During the first round, all models select two random sequences to constitute their initial training set. After the completion of training  on the initial training set, the algorithm selects a single sequence from the pool of available unlabeled sequences $s_i \in S_{pool}$. This is repeated for 13 rounds and associated performance and annotation-costs are recorded for each round. During training, we group the 23 fine-grained classes in \texttt{FOCAL} into 4 superclasses which include pedestrian, bicycle, car, and cart. Additionally, we formed the train, test, and validation based on a rough 70\%/20\%/10\% split of the data. Furthermore, it was ensured that every unique scene belonged to a separate split and that each split contained a variety of weather and lighting conditions.  This entire training process varies slightly depending on whether the query strategy is an inferential or conformal sampling algorithm.

\paragraph{Inferential Sampling} We define traditional active learning algorithms that compute posterior probabilities  to rank samples in the unlabeled pool as inferential approaches. We adapt these algorithms, defined on single frames, to entire video sequences. We evaluate the acquisition function on entire sequences of video data in the form of $s_i = \{x_{i,1}, ..., x_{i,N_{s_i}}\}$. Here $x_{i,k}$ represent the individual frames within sequence $s_{i}$ and $N_{s_i}$ represents sequence length. The acquisition function $a$ is defined as:

\begin{equation}
\label{eq:seq-al}
    S^* = \argmax_{s_1, ..., s_b \in S_{pool}} a(s_1, ..., s_b | g(f_w, S_{pool}))
\end{equation}

where $g$ represents a scoring function that takes in the trained network parameters $f_w$ and the entire pool of unlabeled sequences $S_{pool}$ and produces scores for each sequence. The acquisition function $a$ then selects an optimal $S^*$ based on the associated scores and their corresponding sequences. $g$ and $a$ vary depending on the active learning algorithm of interest. In our implementations (and \texttt{FOCAL} benchmarks), $g$ computes a score for the sequence as a whole by averaging the scores acquired for each frame in the associated sequence.

 To generate the discussed scores, we adapt existing algorithms from a classification to an object detection setting. For entropy, which is based on softmax probabilities \cite{wang2014new}, we use the object probability of each YOLOv5 grid in the output prediction and derive a single acquisition function score for each grid. Each frame score is then reduced to averaging the grid scores within a single frame. Following this intuition, we produce an overall sequence score by averaging all frame scores within a sequence. This process is repeated across all methods that make use of logit probability outputs which includes least confidence, margin, badge, and coreset. For FALSE \cite{benkert2022false} and GauSS \cite{benkert2022gauss} we define the switching score as the difference between the predicted number of objects in between rounds for each frame and produce a sequence score by calculating the arithmetic mean across all frames. We sample the highest switching scores for FALSE while GauSS approximates the switching score distribution with a two component Gaussian mixture model and samples from the component with a higher switch mean. We also perform random sampling where we randomly select the next batch of video sequences from the unlabeled pool.

\paragraph{Conformal Sampling} Inferential strategies are typically model dependent. However, conformal sampling does not require posterior probabilities from a trained model to select samples from the unlabeled pool. Rather, each algorithm selects its initial samples according to its unique criteria. The following are descriptions of the conformal sampling algorithms proposed in this study:

\begin{itemize}
\item Least Frame: This approach selects the sequence with the shortest length in $S_{pool}$. This utilizes a function $l(S_{pool})$ that outputs the length of each sequence. It has the form $S^* = \argmin_{s_{i}} (l(S_{pool}))$

\item Most Frame: This approach selects the sequence with the longest length in $S_{pool}$. It has the form  $S^* = \argmax_{s_{i}} (l(S_{pool}))$

\item Motion: This is a motion-based approach that uses optical flow \cite{fleet2006optical} as a metric to rank sequences according to its temporal information. Optical flow maps for each sequence are generated using a PWC-Net \cite{sun2018pwc} model trained on the KITTI 2015 dataset \cite{geiger2015kitti}. Motion scores, $\Sigma_{i = 0}^{N_{s_j}}m_i(x_{i,s_{j}})$, are generated by summing the pixel values of each flow map generated for every frame $x_{i,s_{j}}$ within a sequence $s_{j}$. Low motion sequences have smaller pixel values in their flow maps and the opposite is true for high motion sequences.  $N_{s_j}$ is the number of frames in sequence $s_{j} \in S_{pool}$. 
    \begin{itemize}
        \item Min Motion: This selects the next sequence with the lowest motion scores
        \begin{equation}
        \label{eq:min-motion}
            S^* = \argmin_{s_{j}}\Sigma_{i = 0}^{N_{s_j}}m_i(x_{i,s_j})
        \end{equation}
        \item Min Max Motion: This selects $k$ sequences having either the lowest or highest motion scores at alternating rounds $r$.

        \[ k^* = \begin{cases} 
              \argmin_{s_{j}}\Sigma_{i = 0}^{N_{s_j}}m_i(x_{i,s_j}) & r \in E \\
              \argmax_{s_{j}}\Sigma_{i = 0}^{N_{s_j}}m_i(x_{i,s_j}) & r \in O \\
           \end{cases}
        \] 
        $E$ and $O$ are the set of even and odd rounds respectively.
    \end{itemize}
\item Min Boxes: This is a sampling approach that uses the spatial distribution of pixels in optical flow maps to estimate the number of boxes $\Sigma_{i = 0}^{N_{s_j}}b_i(x_{i,s_j})$ in a sequence where $b_i(x_{i,s_j})$ is the estimated number of boxes in a frame. This strategy selects the sequence with the lowest estimated number of boxes and has the form:
        \begin{equation}
        \label{eq:min-boxes}
            S^* = \argmin_{s_j} \Sigma_{i = 0}^{N_{s_j}}b_i(x_{i,s_j})
        \end{equation}
\end{itemize}

In Figure \ref{fig:upper-lower-cost-bounds}, we perform an analysis of the annotation-cost of sequences selected by these conformal strategies. We show where annotation cost of each sampling method lies with respect to the theoretical bounds (dashed lines) of \texttt{FOCAL}. Theoretical upper and lower bounds are computed by selecting sequences in descending and ascending cost order, respectively, and then summing cumulatively. For ease of visualization, we show the first seven active learning rounds. Using the estimates of motion in conformal sampling, the red and green curves are closer to the theoretical minimum (dashed light blue). Since cost is less correlated to sequence length, it is not surprising that least and most frame sampling (blue and orange curves) have cumulative costs closer to the upper bound (dashed dark blue). Inferential methods are associated with a stochastic annotation cost and the average and standard error of each is shown as a shaded curve.

\subsection{Performance-Cost Metrics}
Within the context of active learning, cost analysis is especially relevant when considered jointly with generalization performance. For instance, selecting low cost video sequences is less relevant when no additional detection performance gain is expected. For this purpose, we visualize changes in both via mAP versus cost plots. The x-axis represents the cost, in hours, of labeling the number of sequences in the training pool and the y-axis is the mAP the model achieves on the test set. Based on this, we develop two metrics to study performance-cost evaluation: cost appreciation rate (CAR) and performance appreciation rate (PAR). 

\paragraph{Cost Appreciation Rate}
The cost appreciation rate is the area under the mAP vs cost curve at different cost budgets. CAR is defined as follows:

\begin{equation}
\label{eq:car}
    CAR = \int_{0}^{b} AP({c}) \,dc.
\end{equation}

where $b$ is the cost budget, $c$ is a cost value less than the budget, and AP({c}) is the corresponding mAP on the y-axis for cost value $c$. CAR plots shown in Section \ref{sec:results} answer the question, ``With cost-focused budgets leading up to $b$ hours maximum, which algorithm results in the highest performance at each budget?". High CAR values are desirable.

\paragraph{Performance Appreciation Rate}
Similar to CAR, the performance appreciations rate is the area under the mAP vs cost curve at different performance budgets. PAR is defined as follows:

\begin{equation}
\label{eq:par}
    PAR = \int_{0}^{b} AP({p}) \,dp.
\end{equation}

where $b$ is now the performance budget, $p$ is a performance value less than the budget, and $AP({p})$ is the corresponding cost on the x-axis for performance value $p$. PAR plots shown in Section \ref{sec:results} answer the question, ``With budgets leading up to $b$ mAP, which algorithm achieves the lowest cost at each budget?". Low PAR is indicative of better performance.

\paragraph{Overhead Computation Cost}
In addition to annotation-cost, there is also a cost associated with the querying process for sequences. For inferential sampling techniques (except random), overhead cost is defined as the number of floating point operations per second (FLOPS) YOLOv5n executes to compute the posterior probabilities of the unlabeled pool for all active learning rounds. The overhead cost for Min Boxes, Min Motion and Min Max Motion conformal sampling is the FLOPS of PWC-Net to compute the optical flow maps for the available data before beginning active learning. Random, Least Frame and Most Frame sampling have zero overhead cost.
\vspace{-1mm}
\section{Results}
\label{sec:results}

\begin{figure*}[h!]
\tiny
	\centering
	\includegraphics[width=\textwidth]{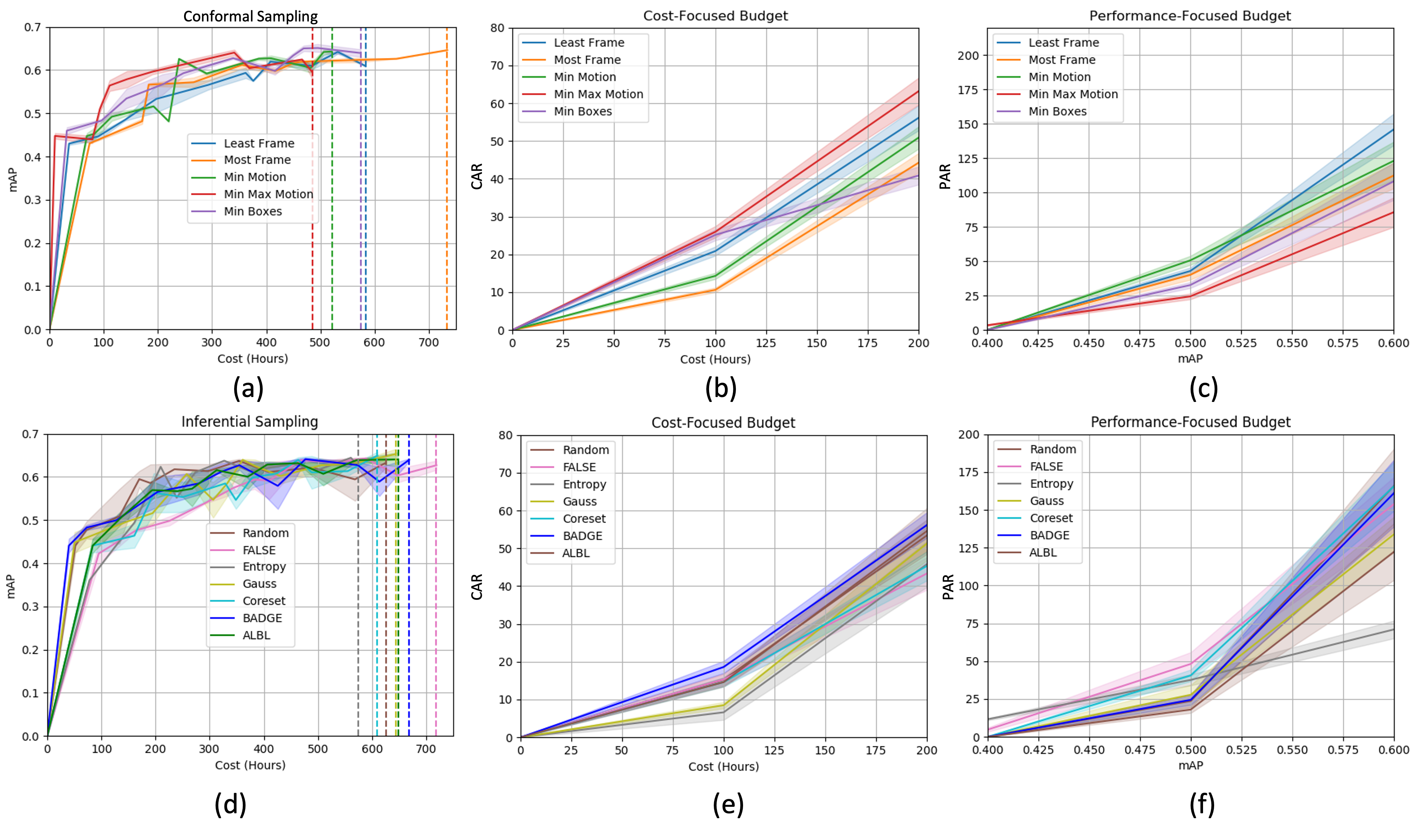}
	
	\caption{A comparison between cost estimated and probabilistic sampling methods for: (a, d) mAP versus cost. (b, e) cost appreciation rate (CAR) versus cost. (c, f) performance appreciation rate (PAR) versus performance.} \vspace{-.4cm}
	
	\label{fig:cost-aware-unaware-cpmpare}
\end{figure*}

\begin{figure}[h!]
\tiny
	\centering
	\includegraphics[width=0.7\columnwidth]{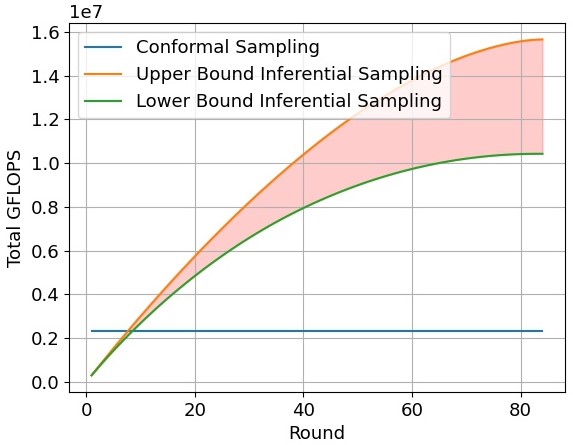}
	
	\caption{A comparison of overhead cost in Giga-FLOPS (GFLOPS) between inferential and conformal sampling} \vspace{-.6cm}
	
	\label{fig:gflops}
\end{figure}

\paragraph{Object Detection} We first benchmark the performance of standard object detection strategies on \texttt{FOCAL} and compare it to performance on the COCO \cite{lin2014microsoft} dataset.
We show the mean average precision (mAP) metric at IOU thresholds $0.5$ (mAP@$50$), and mAP between $0.5$ and $0.95$ (mAP@$[50, 95]$) on \texttt{FOCAL} in the last two columns of Table~\ref{tab:detection_benchmarks}. While YOLOv5n has the lowest number of parameters, it achieves the highest test mAP@$50$ of $0.640$. We also show the detection performance of the same architecture on the COCO dataset as mAP@$50$ (COCO). Note that all  models are trained and tested on the $4$ superclasses of \texttt{FOCAL}, while the same architectures are trained on the $80$ object categories of the COCO with higher class complexity. Thus, the overall detection scores of these architectures are lower on the COCO test-dev \cite{redmon2018yolov3, duan2019centernet, Jocher_YOLOv5_by_Ultralytics_2020}. However, \texttt{FOCAL} exhibits various challenges for object detection, especially in the context of infrastructure-based autonomous vehicle research. The challenges include the variation of object size, motion, occlusion, and location, as shown in Fig~\ref{fig: space_temp_diversity}.
Considering the trade-off between performance and training time, we use YOLOv5n in the following active learning experiments.

\begin{table} [h!]
\small
\centering
\caption{Object Detection Benchmarks on \texttt{FOCAL}.}
\label{tab:detection_benchmarks}
\scalebox{0.8}{
\begin{tabular}{cccc} 
\toprule
\textbf{Architecture} & \textbf{mAP@50 (COCO)}    & \textbf{mAP@50} & \textbf{mAP@[50, 95]}  \\ 
\hline \midrule
\rowcolor{yellow} YOLOv5n &0.460     & 0.640           & 0.396                  \\
YOLOv5m                   &0.639     & 0.620           & 0.388                  \\
YOLOv5x                   &0.689      & 0.628           & 0.385                  \\
YOLOv3                    &0.579        & 0.612           & 0.402                  \\
YOLOv3-tiny               &0.331        & 0.574           & 0.293                  \\
CenterNet                 &0.416         & 0.538           & 0.330                  \\
\bottomrule
\end{tabular}} 
\end{table}

\paragraph{Sequence Active Learning}
The presence of cost labels in \texttt{FOCAL} allows an exploration of the relation between mAP and annotation cost which is shown in Figure \ref{fig:cost-aware-unaware-cpmpare}(a, d). We show changes in mAP and cost for 12 rounds of active learning on three unique random seeds in Figure \ref{fig:cost-aware-unaware-cpmpare}(a, d). Shading around each curve represents the standard deviation error and the colored dashed vertical lines indicate the total cost of labeling all 13 sampled sequences at the end of 12 rounds. Overall, performance is comparable between both types of sampling methods. However, conformal sampling is noticeably cheaper for some strategies. The Min Max Motion sampling (red) in Figure \ref{fig:cost-aware-unaware-cpmpare}(a) has the shortest width overall at just under 500 hours of cost. This indicates that it minimizes cost the most with a gap of 113 hours compared to the closest inferential sampling approach (entropy) while maintaining comparable mAP to all other strategies. Additionally, it is interesting to observe that sequences with the most and least frames follow similar trends in terms of performance while still costing more than the motion based query strategy. These improvements are more pronounced in plots of CAR and PAR compared to annotation-cost. 

We visualize CAR and PAR metrics in Figures \ref{fig:cost-aware-unaware-cpmpare}(b-c, e-f) given budgets in cumulative cost or performance. Specifically, comparing Figures \ref{fig:cost-aware-unaware-cpmpare}(c,f), we see Min Max Motion (red) has the highest CAR overall. This means it is the best at maximizing performance on \texttt{FOCAL} given cost-focused budgets. Conversely, comparing Figures \ref{fig:cost-aware-unaware-cpmpare}(c,f) shows that at a budget of 0.525 mAP, Min Max Motion sampling (red) is the cheapest algorithm overall. However, when the budget is 0.6 mAP Entropy (grey) becomes the cheapest. This stems from the non-linear fashion in which models learn in machine learning. These kinds of analysis are only possible on \texttt{FOCAL} dataset because of the availability of cost labels.  Additionally, these results suggest that query strategies that make use of a factor that influences cost have a better trade-off between cost and performance compared to previous inferential active learning strategies. There are a variety of potential reasons for this behavior. Within an object detection setting, identifying sequences with a wide variety of motion statistics may be more useful for downstream detection performance. Another consideration is that most active learning strategies don't reflect the real-world annotation process. In this case, strategies like min-max motion are enabled by the sequential nature of our experimental setup and as such are better candidates for real-world video active learning paradigms. 

In Figure \ref{fig:gflops} we show the overhead cost on all available \texttt{FOCAL} data during active learning experiments. For inference on a single \texttt{FOCAL} frame, YOLOv5n executes 4.1 GFLOPS. Inference with PWC-Net on a pair of frames to generate an optical flow map utilizes 30.54 GFLOPS. The overhead cost of inferential sampling at a single round is the product of 4.1 GFLOPS and the frame count of the remaining sequences in the unlabeled pool at that round. This is computed for each active learning round and then cumulatively summed to arrive at the total overhead cost. Conversely, the total overhead for conformal sampling is the product of 30.54 GFLOPS and the total frames available for training which results in just over 2.3 million GFLOPS. Red shading shows how overhead for inferential sampling cumulatively increases within the theoretical bounds while it remains constant (blue line) for conformal sampling methods. The reason for this is conformal sampling does not perform inference on the unlabeled pool at the end of every round. Instead, its overhead to compute the flow maps is front loaded and only performed once. The upper and lower theoretical bounds are calculated by selecting sequences in ascending and descending sequence length order respectively multiplying by 4.1 GFLOPS per frame and then cumulatively summing across all frames. \vspace{-2mm}

\section{Conclusion}
In this paper, we introduce \texttt{FOCAL} - the first public cost-aware video active learning dataset for object detection. We show that \texttt{FOCAL} acts as a benchmark for both active learning and object detection tasks. Within the context of active learning, we conduct a detailed cost analysis to understand the factors that induce cost during the human annotation of video sequences. We then exploit these factors to design novel conformal active learning sampling algorithms that better balance cost and performance considerations. Additionally, the prospective real time annotation labels of \texttt{FOCAL} paves an avenue for accurate, practical, cost-oriented video active learning approaches. It allows an in depth performance-cost analysis using mAP versus cost curves along with CAR and PAR metrics. We see \texttt{FOCAL} as a first step towards integrating video active learning in practical deployment scenarios.

\bibliographystyle{IEEEtran}
\bibliography{IEEEexample}
\vspace{12pt}

\end{document}